\documentclass{SPAICE}

\def\authorEmail{stefano.silvestrini@polimi.it}
\def\AuthorShort{S. Silvestrini et al.}

\author[1]{Stefano Silvestrini\thanks{Corresponding author. E-Mail: \authorEmail}}
\author[1]{Michele Ceresoli}

\affil[1]{Politecnico di Milano, Via Giuseppe La Masa, 34, 20156, Milan, Italy}

\title{On the Geometry of Learned Representations in Event-Based Multi-Modal Egomotion Estimation}

\begin{document}

\maketitle

\begin{abstract}
Classical approaches to event-based egomotion estimation, including those adopted by the top-performing teams of the ELOPE challenge, rely on geometric optimization frameworks such as contrast maximization, homography estimation, or dense optical flow combined with analytic motion inversion. This work investigates the geometric structure that emerges inside a multi-modal network for egomotion estimation. Event tensors, inertial measurements, and range signals are fused through a cross-modal attention architecture and trained in a batch setting. We analyze the latent space geometry and attention dynamics, showing that (i) embeddings lie on low-dimensional manifolds aligned with motion variables, (ii) attention weights adapt with angular excitation and visual reliability, and (iii) the fused representation recovers classical observability cues. These results bridge analytical estimation theory and modern data-driven fusion.
\end{abstract}

\section{Introduction}

Egomotion estimation is traditionally formulated through rigid-body kinematics and projective geometry. In frame-based vision, motion estimation is often derived from brightness-constancy assumptions, whereas event-based methods usually exploit temporal contrast and event alignment. A comprehensive overview of the ELOPE dataset, competition design, and submitted approaches is provided in~\cite{elope_challenge}, which serves as the primary reference for contextualizing this work. Contrast maximization techniques \cite{gallego2018unifying} estimate motion parameters by warping events according to a candidate motion hypothesis and maximizing image sharpness or frequency-domain energy. These methods operate directly on asynchronous event streams and adaptively select temporal windows based on event density, enabling high temporal fidelity. Extensions of this paradigm have been applied to lunar landing and space navigation scenarios \cite{mcleod2022globally, silvestrini2026deep}, demonstrating excellent accuracy but requiring iterative optimization at inference time. Deep networks operating on event tensors \cite{zhu2018evflownet,gehrig2021eraft} or fusing event and inertial data \cite{mueggler2018continuous} directly regress motion quantities, shifting complexity from inference to training. This raises a central question: \textit{does geometry disappear in learned fusion models, or is it encoded implicitly in their representations?} This paper addresses this question by analyzing the latent space of a self-supervised multi-modal network for vertical egomotion estimation during simulated lunar descent. We use the novel end-to-end self-supervised model as a learned multi-modal case study to investigate whether geometric structure emerges in its internal representation.
These results do not replace classical geometric estimators, but make explicit how part of their geometric reasoning is amortized inside the learned representation. The analysis suggests practical uses of latent-space diagnostics for reliability monitoring, failure detection, and future hybrid learned--optimization pipelines. The contribution of this work is therefore not a state-of-the-art performance claim, but a diagnostic study of the geometry learned by a newly developed multi-modal event-based model. In particular, we provide:
\begin{itemize}
    \item a state-sufficiency test showing that the instantaneous latent vector contains almost all information needed for velocity prediction;
    \item latent-space reliability diagnostics showing that high-error windows can be detected from embedding norms and Mahalanobis distance;
    \item a physical-alignment analysis showing that dominant latent directions correlate with velocity-related variables;
    \item an interpretation of cross-modal attention as an implicit reliability modulation mechanism.
\end{itemize}
These analyses make the comparison with classical optimization-based methods explicit: while classical methods solve a geometric problem online, the learned model appears to amortize part of this geometric structure into its latent representation.

\section{Model Architecture}
\begin{figure}[t]
  \centering
  \includegraphics[width=.99\columnwidth]{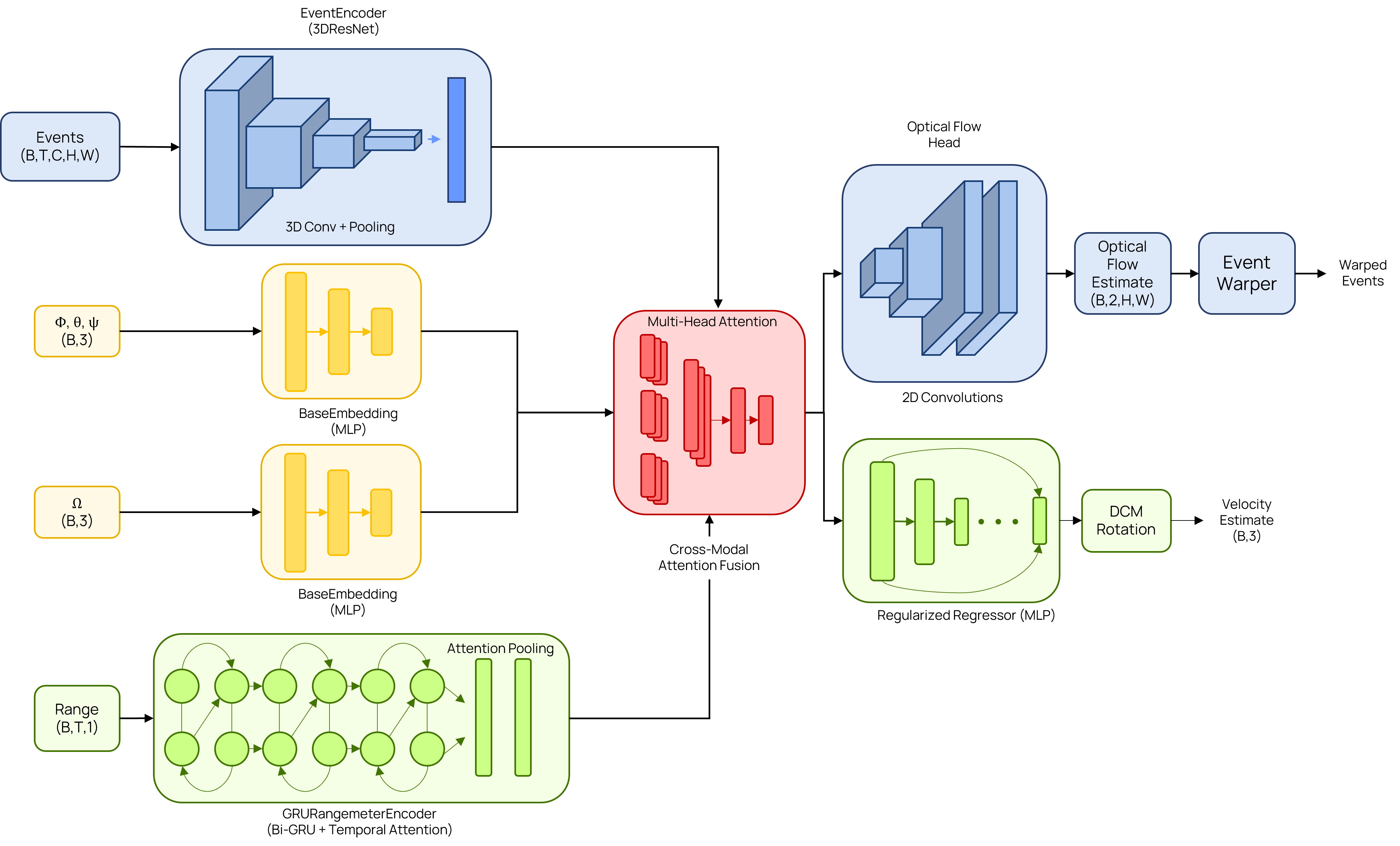}
  \caption{Diagram of \texttt{emmnet} model architecture }
  \label{fig:emmnet-model}
\end{figure}

The model architecture is shown in Fig.~\ref{fig:emmnet-model}.

\subsection{Input Representation}

Each training sample contains a multimodal sequence of length $S$. The model consumes:
\begin{itemize}
\item Event tensor at the final timestep:
$\mathbf{E} \in \mathbb{R}^{B \times S \times 2 \times C \times H \times W}$,
\item IMU Euler angles and angular rates:
$\boldsymbol{\theta}, \boldsymbol{\omega} \in \mathbb{R}^{B \times 3}$,
\item Rangemeter sequence:
$\mathbf{r} \in \mathbb{R}^{B \times S \times 1}$,
\end{itemize}
where $B$ denotes batch size.

\subsection{Sensors Encoders}

The \textbf{event tensor} is processed by a 3D ResNet-style encoder composed of spatio-temporal convolutions and residual blocks. Convolutions jointly operate across the temporal encoding dimension and spatial dimensions, progressively increasing feature depth. After the final stage, adaptive average pooling reduces spatial dimensions to $2\times2$, producing
$
\mathbf{F}_e \in \mathbb{R}^{B \times D_e \times 2 \times 2}.
$
This tensor is reshaped into four event tokens,
$
\mathbf{T}_e \in \mathbb{R}^{B \times 4 \times D_e},
$
which serve as structured visual descriptors for fusion. \textbf{IMU embeddings} from Euler angles and angular velocities are encoded independently by lightweight multilayer perceptrons with LayerNorm and nonlinear activations. This yields embeddings
$
\mathbf{f}_{\theta}, \mathbf{f}_{\omega} \in \mathbb{R}^{B \times D_i}.
$
These embeddings provide compact inertial state descriptors at the final timestep.
Finally, \textbf{rangemeter measurements} are processed across the temporal dimension using a bidirectional GRU. An attention pooling mechanism computes weights over the sequence and produces a fixed-dimensional embedding
$
\mathbf{f}_r \in \mathbb{R}^{B \times D_r}.
$
This design preserves temporal context while compressing the range sequence into a single state vector.

\subsection{Cross-Modal Attention Fusion}

All modality embeddings are projected into a shared latent space of dimension $D$. The four event tokens are concatenated with the IMU and rangemeter embeddings, forming a sequence of seven tokens. Multi-head self-attention is applied over this sequence, followed by a feed-forward block with residual connections and normalization. The resulting representations are aggregated to produce a fused feature vector
\begin{equation}
\mathbf{f} \in \mathbb{R}^{B \times D},
\end{equation}
which encodes cross-modal interactions between visual, inertial, and range information.

\subsection{Velocity Regressor}

The fused representation is processed by a regularized regression head consisting of fully connected layers with LayerNorm, dropout, GELU activations, and a skip connection. The network predicts a body-frame velocity
\begin{equation}
\hat{\mathbf{v}} \in \mathbb{R}^{B \times 3}.
\end{equation}
The predicted body-frame velocity is rotated into the inertial frame using a direction cosine matrix constructed from the Euler angles:
\begin{equation}
\hat{\mathbf{v}}_{\text{inertial}} =
\mathbf{R}(\boldsymbol{\theta}) \hat{\mathbf{v}}.
\end{equation}
This inertial-frame velocity constitutes the final model output.

\subsection{Self-Supervised Optical Flow Head}

An auxiliary optical flow head can be attached to the event features. Given two consecutive event windows $E_0$ and $E_1$, polarity/time channels are flattened to obtain
\begin{equation}
\tilde{E}_t \in \mathbb{R}^{C' \times H \times W}.
\end{equation}
The model predicts a dense flow field
\begin{equation}
F=(u,v) \in \mathbb{R}^{2 \times H \times W}.
\end{equation}
A differentiable warp operator using bilinear sampling is defined in order to derive the photometric loss. The photometric reconstruction loss $\mathcal{L}_{\text{photo}}$ is regularized by a first-order smoothness term $\mathcal{L}_{\text{smooth}}$, yielding:
\begin{equation}
\mathcal{L}_{\text{flow}} =
\lambda_{\text{photo}} \mathcal{L}_{\text{photo}}
+ \lambda_{\text{smooth}} \mathcal{L}_{\text{smooth}},
\end{equation}
Thus, the total training loss is composed of the supervised term $\mathcal{L}_{\text{vel}}$ and the self-supervised task:
\begin{equation}
\mathcal{L}_{\text{total}} =
\mathcal{L}_{\text{vel}} +
\lambda_{\text{aux}} \mathcal{L}_{\text{flow}}.
\end{equation}

\section{Latent-Space Diagnostics and Relation to Classical Estimation}
We analyze the latent representation $z_t$ immediately before the regression head. The following diagnostics investigate whether classical properties of state estimators reappear in representation space. Table~\ref{tab:classical_comparison} clarifies the intended comparison. We do not claim that the learned representation provides the same guarantees as an explicit optimizer. Rather, we test whether quantities that play a central role in classical estimators have measurable analogues in latent space.
\begin{table}[t]
\centering
\caption{Interpretation of the proposed diagnostics in relation to classical optimization-based egomotion pipelines.}
\label{tab:classical_comparison}
\begin{tabular}{p{0.34\columnwidth}p{0.28\columnwidth}p{0.28\columnwidth}}
\toprule
Property & Classical geometric pipeline & Latent-space diagnostic \\
\midrule
Compact motion state & Explicit motion parameters estimated per window & $z_t$ linearly predicts $v_t$ with high $R^2$ \\
Local sensitivity & Objective curvature / residual sensitivity & Mahalanobis distance and latent norm increase in difficult regimes \\
Measurement reliability & Residuals, event contrast, tracking quality & Latent deviation predicts high-error windows with AUC $>0.8$ \\
Sensor weighting & Explicit covariance or heuristic weighting & Attention weights vary with angular excitation and event reliability \\
Inference mechanism & Online iterative optimization & Amortized single forward pass \\
\bottomrule
\end{tabular}
\end{table}
\subsection{State Sufficiency and Instantaneous Geometric Encoding}
In classical geometric pipelines, motion is estimated from a compact set of parameters that summarize the information in a measurement window. Once this geometric solution is obtained, additional historical measurements provide limited incremental benefit for the current estimate. We evaluate this property by regressing velocity $v_t$ from:
\begin{equation}
z_t, \quad [z_t, z_{t-1}], \quad [z_t, z_{t-1}, z_{t-2}]
\end{equation}
The coefficient of determination $R^2$ improves only marginally when adding history. The minimal history gain indicates that $z_t$ behaves as a nearly sufficient state representation. This indicates that the learned embedding behaves as a compact geometric summary of the current measurement window, analogous to the explicit motion parameters extracted by optimization-based pipelines.

\begin{table}[t]
\centering
\caption{Best model latent diagnostics.}
\label{tab:diag}
\begin{tabular}{lc}
\toprule
Metric & Value \\
\midrule
\multicolumn{2}{l}{\textbf{State sufficiency}} \\
$R^2(z_t \rightarrow v_t)$ & 0.995596 \\
$R^2([z_t,z_{t-1}] \rightarrow v_t)$ & 0.996299 \\
$R^2([z_t,z_{t-1},z_{t-2}] \rightarrow v_t)$ & 0.997140 \\
History gain & 0.001544 \\
\midrule
\multicolumn{2}{l}{\textbf{Latent sensitivity and reliability}} \\
AUC high-error detection using $\|z\|$ & 0.826371 \\
AUC high-error detection using $d_{\text{mah}}(z)$ & 0.810946 \\
$\|z\|_{\text{sparse}}/\|z\|_{\text{dense}}$ & 1.016363 \\
$d_{\text{mah}}(z)_{\text{sparse}}/d_{\text{mah}}(z)_{\text{dense}}$ & 1.155766 \\
$\|z\|_{\text{high-speed}}/\|z\|_{\text{low-speed}}$ & 1.267431 \\
$d_{\text{mah}}(z)_{\text{high-speed}}/d_{\text{mah}}(z)_{\text{low-speed}}$ & 2.255575 \\
\midrule
\multicolumn{2}{l}{\textbf{Physical alignment (latent vs kinematics)}} \\
CCA top-1 & 0.989626 \\
CCA top-2 mean & 0.973007 \\
CCA top-3 mean & 0.957579 \\
CCA top-4 mean & 0.850110 \\
\bottomrule
\end{tabular}
\end{table}

\subsection{Latent Sensitivity and Reliability Indicators}
Optimization-based geometric estimators rely on locally smooth objective landscapes, where small changes in motion parameters induce approximately linear variations in measurement alignment \cite{AlainBengio2016,Kornblith2019CKA}. We analyze local neighborhoods in latent space and observe approximately linear transitions between adjacent timesteps. Moreover, Mahalanobis distance in latent space increases significantly during high-speed motion regimes (Table~\ref{tab:diag}). This suggests that the embedding covariance structure is sensitive to excitation intensity. While this is not equivalent to computing the curvature of an explicit geometric objective, it provides an internal diagnostic that varies systematically with motion regime. High-error windows can be detected using latent norms or Mahalanobis distance with AUC above 0.8, indicating that latent geometry encodes reliability information. In other words, by monitoring latent deviation metrics, one obtains a statistically meaningful indication of elevated estimation error for in-distribution test data \cite{GalGhahramani2016,Lakshminarayanan2017}.

\subsection{Alignment with Physical Variables}

Canonical correlation analysis (CCA) between latent vectors and physical variables $[v_x, v_y, v_z, \|v\|]$ reveals strong alignment. The top canonical component reaches correlation $\approx 0.99$, demonstrating that principal latent directions correspond directly to kinematic quantities. This suggests that dominant directions in representation space are strongly associated with physical motion variables. Rather than arbitrary embeddings, the network organizes features along motion-relevant coordinates.

\subsection{Joint Interpretation of Latent Manifold Structure}

\begin{figure*}[t]
    \centering
    \includegraphics[width=.32\textwidth]{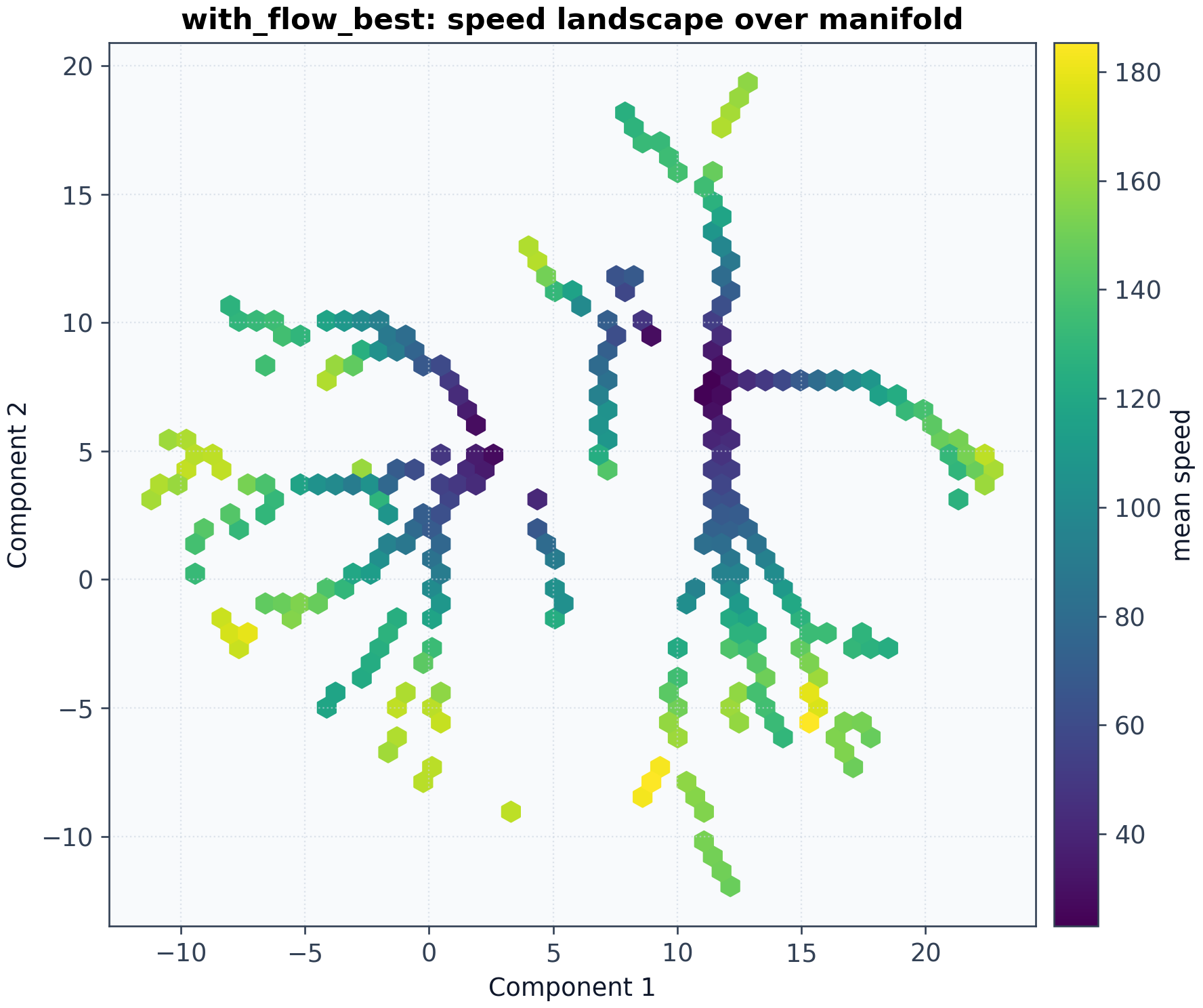}
    \includegraphics[width=.32\textwidth]{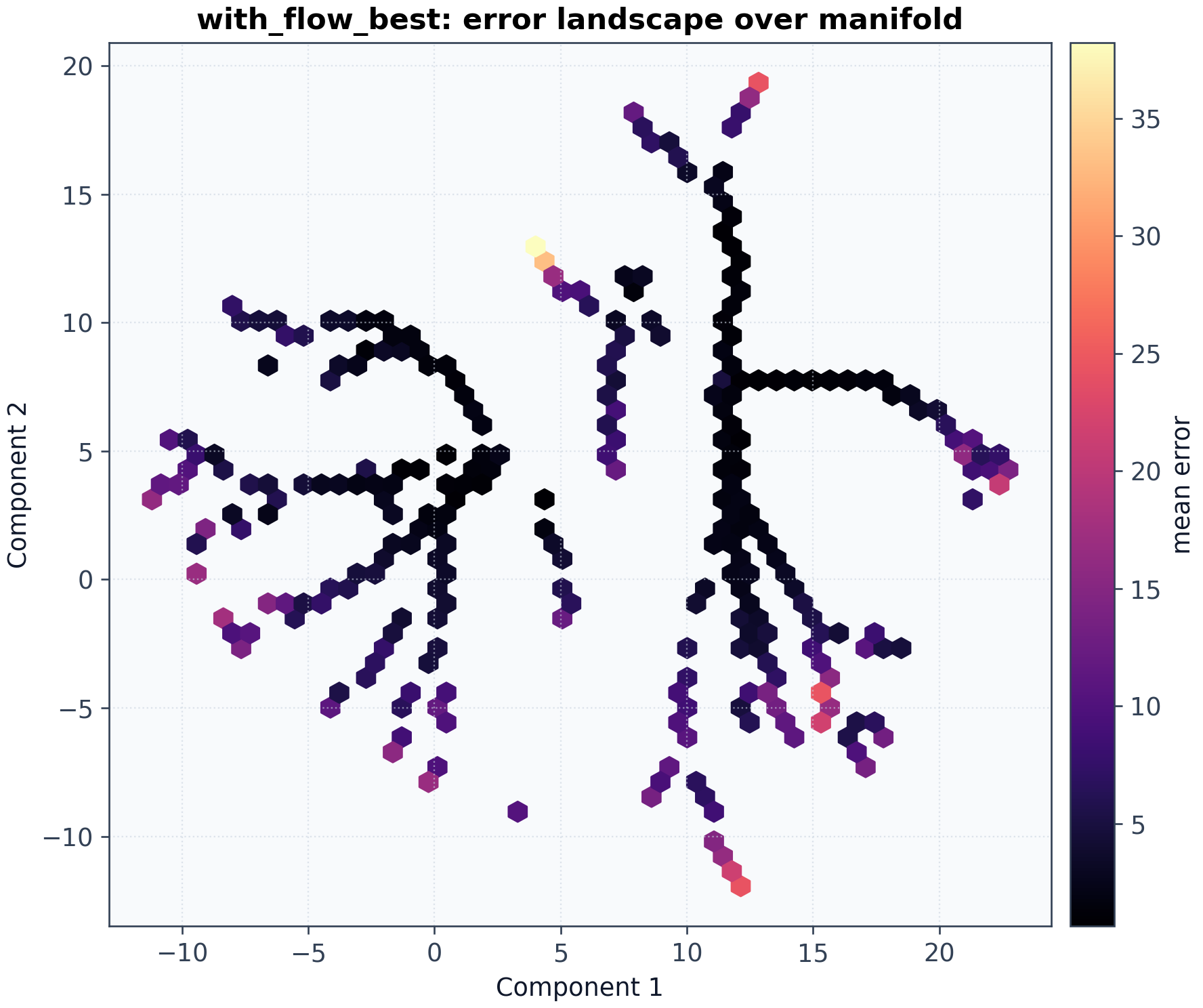}
    \includegraphics[width=.32\textwidth]{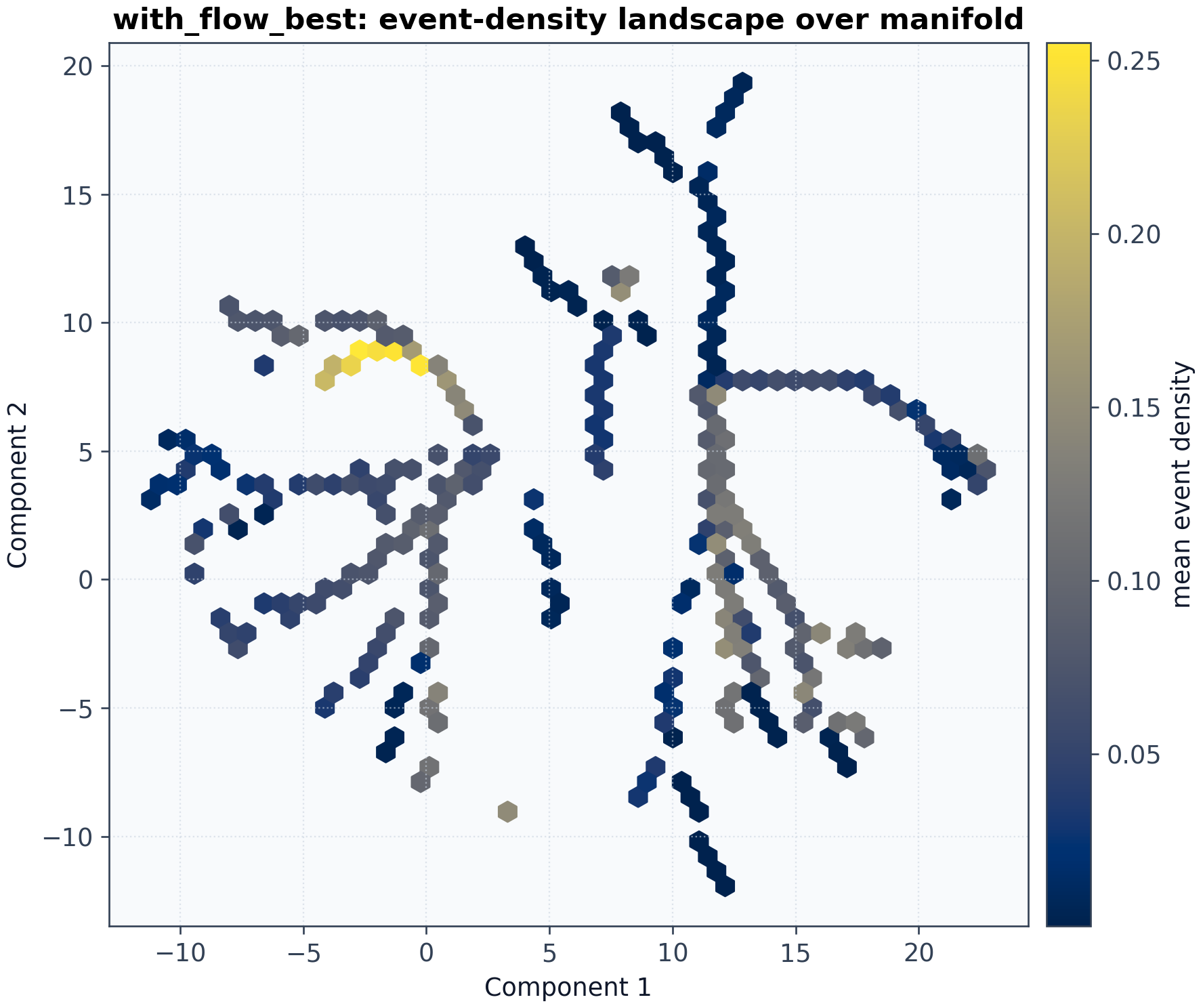}
    \caption{UMAP projection of latent space colored by (left) speed magnitude, (center) prediction error, and (right) event density.}
    \label{fig:umap_joint}
\end{figure*}

Figure~\ref{fig:umap_joint} provides a joint interpretation of speed-structure, event density-structure, and error-structure in the latent manifold. Comparing the three projections reveals that high-error regions are not randomly distributed but coincide with specific motion regimes and excitation densities. Regions of sparse event density occupy structured areas of the embedding, indicating that the representation encodes excitation statistics alongside motion magnitude.

\subsection{Attention as Learned Reliability Modulation}

\begin{figure}[t]
  \centering
  \includegraphics[width=.75\columnwidth]{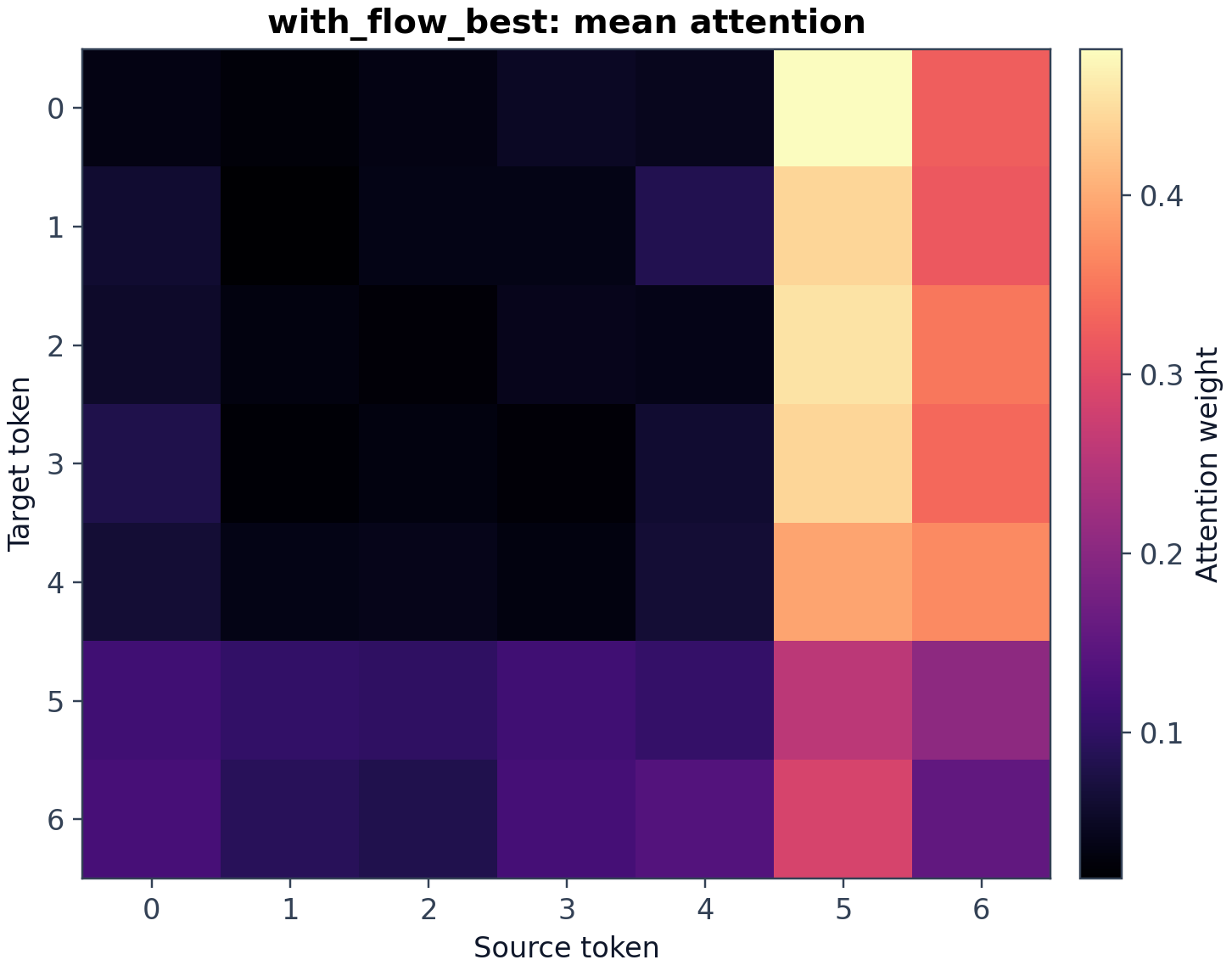}
  \caption{Cross-modal attention weights across descent windows. Attention shifts toward inertial features under strong rotational excitation and toward event features under stable flow conditions. Token order on both axes is \([E_1,E_2,E_3,E_4,\Omega,R,A]\)}
  \label{fig:attention}
\end{figure}

The learned attention pattern is consistent with a dependence on angular velocity magnitude (Fig.~\ref{fig:attention}). During high rotation, the network increases reliance on inertial features, down-weighting event embeddings that become less reliable due to motion-induced degradation or spreading in the event voxel representation. This behavior closely resembles reliability weighting in explicit geometric pipelines: measurement contributions are effectively modulated according to motion regime. Importantly, this modulation is not explicitly programmed—it emerges from training.

\subsection{Practical Implications of the Latent Geometry}

The analyses above suggest three direct uses of the learned representation in space-navigation pipelines.
First, latent-space distances can be used as lightweight reliability monitors. Since both $\|z\|$ and $d_{\mathrm{mah}}(z)$ detect high-error windows with AUC above $0.8$, they can provide an internal warning signal without requiring ground-truth velocity during deployment.
Second, the latent representation can be used to trigger fallback or hybrid estimation strategies. For instance, windows with large Mahalanobis distance or atypical event-density structure could be routed to a slower but more interpretable optimization-based estimator, while nominal windows are processed by the neural model in a single forward pass.
Third, attention weights provide a diagnostic of sensor reliance. A systematic shift toward inertial tokens under high angular excitation suggests that the model learns a form of context-dependent sensor weighting. This can be exploited to identify regimes where event-based visual information becomes less reliable and where inertial or range measurements dominate the estimate.
Therefore, the latent-space analysis is not only descriptive. It points toward monitorable and actionable quantities that could support safety checks in future learned guidance, navigation, and control systems.

\section{Discussion}


The proposed architecture departs from incremental contrast-maximization pipelines by adopting a batch-formulated, amortized inference paradigm \cite{GershmanGoodman2014,KingmaWelling2014}. Instead of explicitly optimizing a geometric objective at each time step, the model learns a mapping from synchronized multi-modal inputs to velocity estimates in a single forward pass. This design reflects two central principles.
\begin{enumerate}
    \item First, multi-modal fusion is performed at the representation level rather than at the output level. Event-derived spatial-temporal features, inertial measurements, and rangemeter signals are projected into a common embedding space and fused through cross-modal attention. This enables the network to learn adaptive weighting of heterogeneous sensing modalities depending on context, rather than relying on fixed analytical fusion rules.
    
    \item Second, supervision is distributed across objectives. Velocity estimation provides explicit geometric supervision, while the auxiliary optical flow head introduces self-supervised constraints that regularize the event encoder. This dual supervision encourages the latent representation to encode both motion-consistent and photometrically coherent structures.
    
    \item The attention-based fusion mechanism was selected not merely for architectural novelty but for its capacity to model cross-modal interactions without imposing hard structural assumptions. In contrast to concatenation-based fusion, attention allows the model to modulate the influence of inertial or range cues depending on event-derived features, which is particularly relevant under challenging illumination or low-texture conditions.
\end{enumerate}
\subsection{Geometry of the Learned Representation}

A central question motivating this work is whether the learned intermediate representation captures geometric structure comparable to explicit state estimators.

Although velocity is supervised at the output layer, the intermediate embedding produced by multi-modal fusion remains unconstrained at the component level and therefore constitutes a latent representation. Empirical analyses indicate that this representation encodes motion-related information in a structured manner:
\begin{itemize}
    \item Linear probing reveals that velocity components can be recovered with high fidelity from the embedding, suggesting approximate linear encoding of motion.
    \item Temporal smoothness of the embedding trajectory correlates with physical motion continuity, indicating structured dynamics in latent space.
    \item Dimensionality analysis shows that the effective intrinsic dimensionality is significantly lower than the embedding dimension, suggesting that the model organizes geometric information along a low-dimensional manifold.
\end{itemize}
These observations support the hypothesis that the network learns an implicit geometric representation rather than an arbitrary feature encoding.

\subsection{Implicit versus Explicit Geometry}

Despite these structured properties, differences between implicit and explicit geometry become apparent when comparing performance against classical pipelines. Classical methods explicitly parameterize motion variables and enforce geometric consistency through optimization. The learned model, by contrast, encodes geometry implicitly in a distributed embedding. While this allows flexibility and amortized inference, it may limit global consistency and long-term metric stability, particularly under distribution shift. In particular, optimization-based methods can re-estimate motion parameters for each measurement window by directly maximizing a geometric consistency objective, whereas the learned representation must rely solely on patterns internalized during training. This may explain scenarios where classical pipelines outperform the learned approach despite the latter’s representational flexibility.

\subsection{Role of Self-Supervision}

The auxiliary optical flow objective plays a regularizing role by encouraging spatially coherent motion representations in the event encoder. Rather than serving solely as an auxiliary prediction task, it shapes the geometry of the latent space by enforcing motion-consistent feature organization. This multi-task structure highlights a broader principle: self-supervised objectives can act as geometric priors when explicit geometric constraints are absent. The interplay between supervised velocity regression and self-supervised flow estimation contributes to the emergence of structured embeddings.

\subsection{Limitations}

The proposed analysis is diagnostic and does not establish formal observability or stability guarantees. The identified latent-space structure is measured empirically on simulated lunar descent data and should therefore be interpreted as evidence of learned geometric organization within the tested distribution. Moreover, although latent distance and attention statistics provide useful reliability indicators, they are not calibrated uncertainty estimates. Future work should evaluate these diagnostics under stronger distribution shifts, sensor degradation, and closed-loop guidance conditions. Overall, the results suggest that learned event-based egomotion estimators do not simply replace geometry with an opaque regression map. Instead, part of the geometric structure used by classical pipelines appears to be amortized into the latent representation: instantaneous embeddings summarize the measurement window, dominant latent directions align with physical velocity variables, and latent deviations correlate with estimation error. The practical value of this observation is that the latent space can be monitored. In future hybrid systems, such diagnostics could be used to trigger fallback optimization, adapt sensor weighting, or reject unreliable estimates before they affect downstream guidance and control.

\subsection{Data \& Code}
The dataset used for training the models can be found at \url{https://kelvins.esa.int/elope/}. The open-source repository of the code is available at \url{https://github.com/stesilve93/elope-apelle}.
\printbibliography
\addcontentsline{toc}{section}{References}

\end{document}